\documentclass[runningheads]{llncs}
\usepackage{graphicx}
\usepackage{hyperref}

\usepackage{amsmath,graphicx}
\usepackage{diagbox}
\usepackage{ textcomp }
\usepackage{mathtools}
\usepackage{bm}
\usepackage{esvect}
\usepackage{url}
\usepackage{pdfpages}
\usepackage{floatrow}
\usepackage{wrapfig}

\usepackage{subcaption}
\usepackage{color}
\usepackage{amssymb}
\usepackage{ dsfont }
\usepackage{afterpage}
\usepackage{balance}
\usepackage{ textcomp }
\usepackage[ruled]{algorithm2e}
\newcommand{\SNAS}{\texttt{SL}$_{\texttt{NAS}}$}
\newcommand{\RBF}{\texttt{RBF}}
\newcommand{\KRIG}{\texttt{KRIG}}
\newcommand{\NAV}{\texttt{NS}}
\newcommand{\DIP}{\texttt{DIP}}
\newcommand{\NN}{\texttt{NN}}
\newcommand{\NAS}{\texttt{NAS}}
\newcommand{\RSS}{\texttt{RSS}}
\newcommand{\IOT}{\texttt{IoT}}
\newcommand{\RSSI}{\texttt{RSSI}}
\newcommand{\MAE}{\texttt{MAE}}
\newcommand{\TV}{\texttt{TV}}

\newcommand{\DR}{\texttt{DR}}

\usepackage{tikz,pgfplots}
\usepackage{pgfplotstable}
\pgfplotsset{compat=1.8}
\usetikzlibrary{pgfplots.statistics}

\DeclareMathOperator*{\argmin}{arg\,min}

\newfloatcommand{capbtabbox}{table}[][\FBwidth]
\begin{document}
\title{Self-Learning for Received Signal Strength Map Reconstruction with Neural Architecture Search}%\thanks{Supported by organization x.}}
\titlerunning{Self-Learning for \RSS{} Map Reconstruction with \NAS}
\author{Aleksandra Malkova\inst{1,2} \and Lo\"{i}c Pauletto\inst{1,3} \and Christophe Villien\inst{2} \and Beno\^{i}t Denis\inst{2} \and Massih-Reza Amini\inst{1}}

\institute{Université Grenoble Alpes, LIG-APTIKAL, F-38401 Saint Martin d'Hères, France \and
Université Grenoble Alpes, CEA-Leti, F-38000 Grenoble, France
 \and Bull/Atos, F-38000 Grenoble, France}

%\orcidID{2222--3333-4444-5555}
\authorrunning{A. Malkova, L. Pauletto et al.}
\maketitle              % typeset the header of the contribution
\begin{abstract}
In this paper, we present a Neural Network (\NN) model based on Neural Architecture Search (\NAS) and self-learning for received signal strength (\RSS) map reconstruction out of sparse single-snapshot input measurements, in the case where data-augmentation by side deterministic simulations cannot be performed. The approach first finds an optimal \NN{} architecture and simultaneously train the deduced model over some ground-truth measurements of a given (\RSS) map. These ground-truth measurements along with the predictions of the model over a set of randomly chosen points are then used to train a second \NN{} model having the same architecture. Experimental results show that signal predictions of this second model outperforms non-learning based interpolation state-of-the-art techniques and {\NN} models with no architecture search on five large-scale maps of \RSS{} measurements.

\keywords{Neural Architecture Search \and Self-learning \and Received Signal Strength \and Radio Mapping}
\end{abstract}
\section{Introduction}
\label{sec:intro}

%The availability of both sensor and radio integrated chips at low cost, and their embedding 
The integration of low-cost sensor and radio chips in a plurality of connected objects %devices, connected objects or 
in the Internet of Things (\IOT{}) %have been widespread in recent years, %As a matter of fact, we are we  witnessing 
has been contributing to the fast development of large-scale physical monitoring and crowdsensing systems in various kinds of smart environments (e.g., smart cities, smart homes, smart transportations, etc). In this context, the ability to associate accurate location information with the sensor data collected on the field opens appealing perspectives in terms of both location-enabled applications and services~\cite{localization_iot_survey}. 

Among possible localization technologies, Global Positioning Systems (\texttt{GPS}) %with centimeters-level accuracy in open sky conditions 
have been widely used in outdoor environments for the last past decades. However, these systems  still suffer from high power consumption, which is hardly compliant with the targeted \IOT{} applications. %which mainly rely on low consumption communication between their elementary nodes (e.g., sensors, actuators).

In order to preserve both nodes' low complexity and fairly good localization performances, an alternative is to interpret radio measurements, such as the Received Signal Strength Indicator (\RSSI) (i.e., received power of sensor data packets sent by \IOT{} nodes and collected at their serving base station(s)), as location-dependent fingerprints for indicating the positions of mobile devices \cite{burghal2020comprehensive,Cheng12,book_wireless_sensor_networks,Tahat16,Yu09}. Typical fingerprinting methods applied to wireless localization~\cite{Vo16} ideally require the prior knowledge of a complete map %describing metrics values (e.g. \RSSI{} values from Gateways) 
of such radio metrics, %for each point belonging to 
covering the area of interest. However, in real life systems, it is
impractical %and even most often prohibitive 
to collect measurements from every single location of the map and one must usually rely uniquely on sparse and non-uniform field data. In order to overcome this problem, %one can estimate the \RSSI{} associated  of each point of a map based on a small set of measurements 
classical map interpolation techniques, such as radial basis functions (\RBF{}) or kriging \cite{LoRa-fingerprint_outdoor},  have been used in this context. These approaches are simple and fast, but they are quite weak in predicting the complex and heterogeneous spatial patterns usually observed in real life radio signals (e.g., sudden and/or highly localized transient variations in the received signal due to specific environmental effects). Data augmentation techniques have thus been proposed for artificially increasing the number of measurements in such radio map reconstruction problems. Typically, once calibrated over a few real field measurements, deterministic prediction tools can generally simulate electromagnetic interactions of transmitted radio waves within the environment of study~\cite{Laaraiedh1,Raspopoulos2,Sorour2}.
The purpose is then to use the generated synthetic data as additional data to train complex models for map interpolation. However, these tools require a very detailed description of the physical environment 
%\textcolor{red}{[BD: We could make it even more concrete to really highlight the practical constraints] "(i.e., physical shapes and constituting materials or dielectric properties of all the objects, buildings... belonging to the scene)"} 
and can hardly anticipate on its dynamic changes over time. Their high computational complexity is also a major bottleneck.

In this paper, we consider \RSSI{} map reconstruction in a constrained low-cost and low-complexity \IOT{} context, where one can rely only on few ground-truth (i.e., \texttt{GPS}-tagged) single-snapshot field measurements and for which data-augmentation techniques based on side deterministic simulations cannot be applied, due to their prohibitive computational cost and/or to \textit{a prior} unknown environment physical characteristics. This problem of map interpolation is similar to the task of image restoration for which, {\NN} based models with  fixed architectures have been already proposed \cite{dip}. In the case where there are few observed pixels in an image these approaches fail to capture its underlying structure that is often complex. To tackle this point we propose a first {\NN} model based on Neural Architecture Search (\NAS) for the design of the most appropriate model given a \RSSI{} map with a small number of ground truth measurements. For this purpose, we develop two strategies based on genetic algorithms and dynamic routing for the search phase. We show that with the latter approach, it is possible to learn the model parameters while simultaneously searching the architecture. Ultimately, in order to enhance the model's predictions, the proposed approach uses also some extra data of the map with the predictions of the optimized {\NN} in non-visited positions together with the initial set of ground-truth measurements for learning a final model.
The proposed technique thus aims at finding practical trade-offs between agnostic learning interpolation techniques and data-augmented learning approaches based on deterministic prediction tools that generally require a very detailed physical characterization of the operating environment. Experimental results on five large-scale {\RSSI} maps show that our approach outperforms non-learning based interpolation state-of-the-art techniques and {\NN} models with a given fixed architecture.
% Considering a fine-tuned \RBF{} interpolation algorithm as a reference baseline, we first aim at finding an optimal by metric Neural Network (\NN{}) model through Neural Architecture Search (\NAS{}). 
% This \NN{} model 

% We first aim at finding an optimal by metric Neural Network (\NN{}) model through Neural Architecture Search (\NAS{}) which is then trained over a set of labeled data with ground truth measurements and predicts signal measures for a set of randomly chosen unlabeled data. These pseudo-labeled examples are added to the set of labeled training examples and another \NN{} model with the same architecture is trained once again over the augmented training set.

\section{Related work}
\label{sec:related_work}
In this section we report related-work on \RSSI{} map reconstruction, as well as existing techniques proposed for \NAS{}.

%\subsection{Wireless communication}
\subsection{Interpolation techniques}
%
%\smallskip
%\noindent \textbf{Agnostic learning interpolation techniques.}
Various spatial interpolation methods have been proposed for radio map reconstruction in the wireless context.

\smallskip

%Perhaps, 
One first approach, known as kriging or Gaussian process regression~\cite{Li_heap_overview_interp}, exploits the distance information between measured points, while trying to capture their spatial dependencies. 
% However, the resulting map does not pass necessarily through the measurement points, contrarily to our problem (imposed ground-truth). 
%
Another popular method is based on radial basis functions (\RBF{})~\cite{LoRa-fingerprint_outdoor,radio_map_interpol_rbf,rbf_interp}. This technique is somehow more flexible, makes fewer assumptions regarding the input data (i.e., considering only the dependency on the distance) and is shown to be more tolerant to some uncertainty \cite{rbf_kriging}. In \cite{LoRa-fingerprint_outdoor} for instance, the authors have divided all the points of a database of outdoor \RSSI{} measurements into training and testing subsets, and compared different kernel functions for the interpolation. The two methods above, which rely on underlying statistical properties of the input data (i.e., spatial correlations) and kernel techniques, require a significant amount of input data to provide accurate interpolation results. Accordingly, they are particularly sensitive to sparse initial datasets. They have thus been considered in combination with crowdsensing. In~\cite{kriging_plus_crowdsensing} for instance, so as to improve the performance of basic kriging, one calls for visiting new positions/cells where the interpolated value is still presumably inaccurate. A quite similar crowdsensing method has also been applied in~\cite{SVT_RSSI_crowdsensing} after stating the problem as a matrix completion problem using singular value thresholding. In our case though, we can just rely on a \RSSI{} map with few ground truth initial measurements.
\smallskip

Another approach considered in the context of indoor wireless localization relies on both collected field data and an a priori path loss model that accounts for the effect of walls attenuation between the transmitter and the receiver~\cite{path_loss_wifi_indoor}. 
%by %using the wall matrix, which 
%counting the number of walls along the path from the access point to the mobile location %and penalize value according to that number, along with 
In outdoor environments, local path loss models (and hence, particularized \RSSI{} distributions) have been used to catch small-scale %wireless topology 
effects in clusters of measured neighbouring points, instead of using raw \RSSI{} data~\cite{path_loss_rssi_outdoor_clustering}. %, where in each cluster the appropriate RSSI distribution model is built. 
However, those parametric path loss models are usually quite inaccurate and require impractical in-site (self-)calibration.
%\textcolor{red}{[BD: What are the main limitations identified with respect to the pb you state?]} 

\smallskip
\noindent\textbf{Data-augmentation approaches.} 
One more way to build or complete radio databases %still 
%envisaged in the context of fingerprinting based positioning  
%consists in relying 
relies on deterministic simulation means, such as Ray-Tracing tools %(e.g., \cite{Raspopoulos1, Raspopoulos2, Sorour1, Sorour2,Laaraiedh1}). 
(e.g., \cite{Laaraiedh1,Raspopoulos2,Sorour2}).
The latter aim at predicting in-site radio propagation  (i.e., simulating electromagnetic interactions of transmitted radio waves within an environment). Once calibrated with a few real field measurements, such simulation data can relax initial metrology and deployment efforts  (i.e., the number of required field measurements). %to build an exploitable radio map. %, or even mitigate practical effects that may be harmful to positioning, such as the cross-device dispersion of radio characteristics (typicaly, between devices used for offline radio map calibration and that used for online positioning). 
Nevertheless, these tools require a very detailed description of the physical environment (e.g., shape, constituting materials and dielectric properties of obstacles, walls...). Moreover, they %are notorious for requiring 
usually require high %and likely prohibitive  
computational complexity. 
%in real applications.
Finally, simulations must be re-run again, likely from scratch, each time minor changes are introduced in the environment.
%that can hardly anticipate on the impact of human activity (like changing  crowd  density, temporary radio link obstructions). 
%
\subsection{\NN{} based models trained after data-augmentation }
%

%\smallskip
%\noindent\textbf{\NN{} based models trained after data-augmentation.}  
Machine and deep learning approaches have been recently applied for \RSSI{} Map reconstruction. These methods have shown to be able to retrieve unseen spatial patterns with highly localized topological effects and hidden correlations. Until now, 
% \textcolor{red}{[BD: We may have missed one relevant contribution based on real measurements, so I would mitigate a bit the statement here] "most of these methods" ?} 
these methods have been trained over simulated datasets given by data-augmentation approaches.
%, , even though relying on quite different settings,.

\smallskip
%A deep learning approach for simulating radio maps is proposed, which takes into account city geography, Tx location, and optionally pathloss measurements and car positions. We devised methods for applying what we knew from the huge dataset of coarsely simulated radio maps to real-world applications. 
In~\cite{RadioUNet}, given a urban environment, the authors introduce a deep neural network called RadioUNet, which outputs radio path loss estimates trained on a large set of generated using the Dominant Path Model data and UNet architecture \cite{RFB15a}.
% \textcolor{red}{[BD: I'm not sure this description is sufficiently self-contained. You may just explain -in brief- those synthetic data, the main principle and main limitations (within 1 or 2 sentences at most).]}]. %, following a supervised learning setting. \textcolor{green}{probably we can remove this commented part? because the work is about another type of the signal}Another work investigates power spectral map estimation using a generative adversarial network (GAN)~\cite{MEGAN}, %called MEGAN~\cite{MEGAN}  which is also trained in a supervised way. This algorithm has been further improved in an unsupervised setting, following a GAN-based pixel regression approach~\cite{MEGAN_PRF}. To estimate the power spectrum maps in the underlay cognitive radio networks, %a two-phase transfer learning-based  a reconstruction algorithm based on a two-phase transfer learning GAN (TPRL-GAN) has been proposed in~\cite{transfer_gan_psm}. 
%A two-phase transfer learning GAN (TPRL-GAN) consists of two stages. First, it projects the source domain into an adjacent domain (domain projecting (DP) framework). Then, extracted features from the adjacent domain are used to reconstruct the full map in the target domain (domain completing (DC) framework). For training the DP, %framework, %the authors also used complete signal distribution maps have been used. 
In another contribution, the authors have shown that using the feedforward neural network for path loss modelling could improve the kriging performance~\cite{kriging_ffnn_pl}, as conventional parametric path loss models admit a small number of parameters and do not necessarily account for shadowing besides average power attenuation. %\textcolor{red}{[BD: I must stop here, I'll resume later tonight.]}

\smallskip

Besides wireless applications, similar problems of map restoration also exist in other domains.  In~\cite{CEDGAN} for instance, the authors try to build topographic maps of mountain areas out of sparse measurements of the altitudes. For this purpose, they use a Generative Adversarial Network (\texttt{GAN}) architecture, where in the discriminator they compare pairs of the input data and the so-called ``received'' map, either generated by the generator or based on the full true map.
%
%That is why machine and deep learning approaches are applicable and seem particularly relevant in this context. 
Another close problem %that is  
%close enough to ours and
making extensive use of neural networks is the image inpainting problem, where one needs to restore missing pixels in a single partial image. By analogy, this kind of framework could be applied in our context too, by considering the radio map as an image, where each pixel corresponds to the \RSSI{} level for a given node location. 
Usually, such image inpainting problems can be solved by minimizing a loss between true and predicted pixels, where the former are artificially and uniformly removed from the initial image. This is however impossible in our case, as only a few ground-truth field measurements %collected on the field 
can be used. %to reconstruct the entire map. 

\smallskip

In contrast to the previous approaches, we consider practical situations where data-augmentation techniques cannot be used mainly because of unknown environment characteristics and computational limitations, and, where there is only a small amount of ground-truth measurements. Our approach automatically searches an optimized Neural Network model for the \RSSI{} map reconstruction in hand, and, it is based on self-training for learning an enhanced \NN{} model with the initial ground-truth and pseudo-labeled measurements obtained from the predictions of the first \NN{} model on a set of randomly chosen points in the map.

\subsection{\NAS{} related methods}
Studies on the subject of \NAS{} have gained significant interest in the last few years. In the literature; there are various of techniques based on Reinforcement Learning (RL) \cite{nas_rl}, evolutionary algorithm \cite{regu_evo} or Bayesian Optimization (BO) \cite{nas_bo}.
Recently, new gradient-based methods  became increasingly popular. One of the first methods using this technique is called DARTS \cite{darts}, in which a relaxation is used to simultaneously optimize the structure of a \textit{cell}, and the weight of the operations relative to each \textit{cell}.  At the end, cells are manually stacked to form a neural network. Based on DARTS, more complex methods have emerged such as AutoDeepLab \cite{autodeep} in which a network is optimized at 3 levels : (i) the parameters of the operations, (ii) the cell structure and (iii) the macro-structure of the network that is stacked manually. Despite a complex representation leading to powerful architectures, this technique has certain drawbacks, such as the fact that the generated architecture is single-path, which means it does not fully exploit the representation's capabilities.  Moreover, as the search phase is over a fixed architecture, it might not be the same between different runs, thus it is complicated to use transfer learning and the impact of training from scratch can be significant. To overcome these limitations, one technique is to use \textit{Dynamic Routing} (\DR) as proposed in \cite{li2020learning}. This approach is different from the traditional gradient based methods proposed for \NAS{} in the sense that it does not look for a specific fixed architecture but generates a dynamic path in a mesh of cells on the fly \textit{without searching}.

\section{NAS for RSSI map reconstruction}

In this section, we first introduce our notations and setting, and then present our main approach, denoted as \SNAS{} in the following.

\subsection{Notations and Setting}

For a given base station $X$, let $Y \in \mathbb{R}^{H \times W}$ be the whole matrix of ground-truth signal measurements
% \textcolor{red}{distribution [BD: right wording ?]}
, where $H \times W$ is the size of the (discretized) area of interest. We suppose to have access to only some ground truth measurements $Y_{m}$ in $Y$, that is  $Y_{m} =~Y \odot M$, where $M \in \{0, 1\}^{H \times W}$ is a binary mask indicating whether each pixel includes one available measurement or not, and $\odot$ is the Hadamard’s product. Here we suppose that the number of non-null elements in $Y_{m}$ is much lower than $H\times W$. We further decompose the measurements set $Y_{m}$ into three parts $Y_{\ell}$ (for \emph{training}), $Y_{v}$ (for \emph{validation}) and $Y_{t}$ (for \emph{test}), such that $Y_{\ell}\oplus Y_{v}\oplus Y_{t}=~Y_{m}$, where~$\oplus$ is the matrix addition operation. Let $X_{\ell}, X_{v}, X_{t}, X_{m}$ be the associated 2D node locations (or equivalently, the cell/pixel coordinates) with respect to base station $X$ and $X_{u}$ be the set of 2D locations for which no measurements are available. 

Our approach is based on three main phases $i)$  \emph{architecture~search~phase} - the search of an optimal architecture of a Neural Network model; $ii)$ \emph{data-augmentation phase} - the assignment of  pseudo-labels to randomly chosen unlabeled data using the predictions of the found \NN{} model trained over $Y_\ell$; and $iii)$ \emph{self-learning~phase} - the training of a second \NN{} model with the same architecture  over the set of initial ground truth measurements and the pseudo-labeled examples. In the next sections, we present these phases in more detail. These phases are resumed in Algorithm \ref{algo:nas-transductive}.

\begin{algorithm}[t!]    
%\scriptsize
\caption{\SNAS}
     \hspace{-5mm}\textbf{Input:} A training set: $(X_\ell,Y_\ell)$; a validation set: $(X_v,Y_v)$ and a set of $2D$ locations without measurements: $X_u$;\\

\hspace{-5mm}\textbf{Init:} Using $(X_\ell,Y_\ell)\cup (X_v,Y_v)$, find interpolated measurements $\tilde Y_u$ over $X_u$ using the \RBF{} interpolation method; \\
\textbf{Step 1:} Search the optimal \NN{} architecture 
% with an evolutionary algorithm
using $(X_\ell,Y_\ell)\cup(X_u,\tilde Y_u)$;\\ 
\textbf{Step 2:} Find the parameters $\theta^\star_1$ of the \NN{} model $f_\theta$~:
\begin{center}
$\theta^\star_1=\mathop{\argmin}_\theta \mathcal{L}(X_\ell,Y_\ell,\theta)$ ~~~\# (Eq. \ref{eq:Loss1});\\
\end{center}
\vspace{-3mm}\textbf{Step 3:} Choose  $X_u^{(k)}$ randomly  from $X_u$
% \textcolor{red}{[BD: $X$ or $X_u$]} 
and find the new parameters $\theta^\star_2$ of the \NN{} model $f_\theta$~:
$\theta^\star_2=\mathop{\argmin}_\theta \mathcal{L}(X_\ell\cup X_u^{(k)},Y_\ell\cup f_{\theta^\star_1}(X_u^{(k)}),\theta)$; \\

\hspace{-5mm}\textbf{Output:} $f_{\theta^\star_2}$,$\tilde Y_u$ 
    \label{algo:nas-transductive}
\end{algorithm}

\subsection{Architecture Search phase}
\label{sec:ASP}
Here, we consider a first reference \RSSI{} map as an input image, where unknown measurements in $X_{u}$ are obtained with a \RBF{} using points in the train and validation sets; $(X_{\ell},Y_{\ell})\cup (X_v,Y_v)$. The latter was found the most effective among other state-of-the-art interpolation techniques \cite{LoRa-fingerprint_outdoor}. Let $\tilde{Y}_{u}$ be the set of interpolated measurements given by \RBF{} over $X_{u}$.  For the search phase of the \NAS{} we have employed two strategies described below. 

\begin{wrapfigure}[18]{r}{0.5\textwidth}
    % \centering
    % \includegraphics[width =0.8 \linewidth]{images/architecture-from-nas_2lines.png}
    \begin{center}
       \vspace{-9mm} \includegraphics[width=0.9 \linewidth]{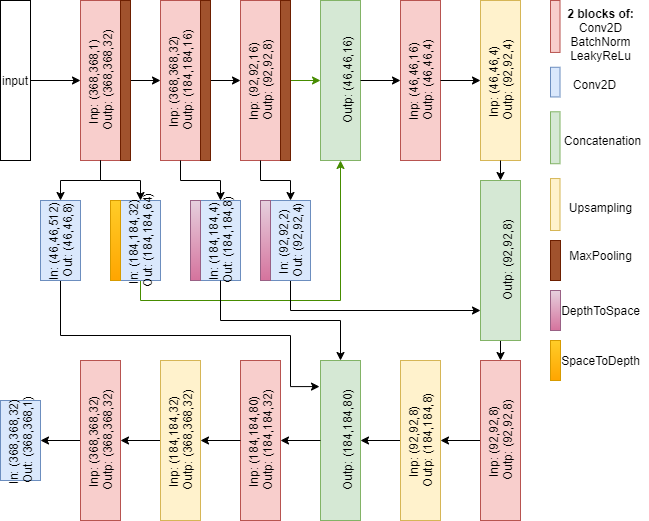}
    \end{center}
\caption{Example of the Neural network architecture found by the Architecture Search phase for the \RSSI{} Map of the city of Grenoble used in our experiments. }
    \label{fig:architecture-nas}
\end{wrapfigure}
\subsubsection{Genetic Algorithm} (\texttt{GA}) 
\label{GA}
From the  set $(X_{\ell},Y_{\ell})\cup (X_{u},\tilde{Y}_{u})$,  we use an evolutionary algorithm similar to \cite{regu_evo} for searching the most efficient architecture represented as a Direct Acyclic Graph (\texttt{DAG}). Here, the validation set $(X_v,Y_v)$ is put aside for hyperparameter tuning. 
The edges of this DAG represent data flow with only one input for each node, %in the graph 
which is a single operation chosen among a set of candidate operations. We consider  usual operations in the image processing field; that are a mixture of convolutional and pooling layers. We also consider three variants of 2D convolutional layers as in \cite{dip} with kernels of size 
% 4, 8 and 64
3, 5 and 7; and two types of pooling layers that compute either the average or the maximum on the filter of size 4. Candidate architectures are then built from randomly sampled operations and the corresponding \NN{} models are trained. The $30$ resulting architectures are then %randomly sampled and 
ranked according to a pixel-wise Mean Absolute Error (\MAE{}) criterion between the interpolated result of the network and the set of interpolated measurements given by \RBF{} $\tilde{Y}_{u}$. The most performing one is finally selected for mutation and placed in the trained population. The oldest architecture is removed in order to keep the size of the population equal to $20$ as in \cite{regu_evo}. Figure \ref{fig:architecture-nas} illustrates such an optimized architecture with 18 nodes, which was found for the \RSSI{} Map of the city of Grenoble used in our experiments (Section \ref{subsec:datasets}).

\subsubsection{Dynamic Routing} 
\begin{figure}[h]
    \centering
    \includegraphics[scale=.35]{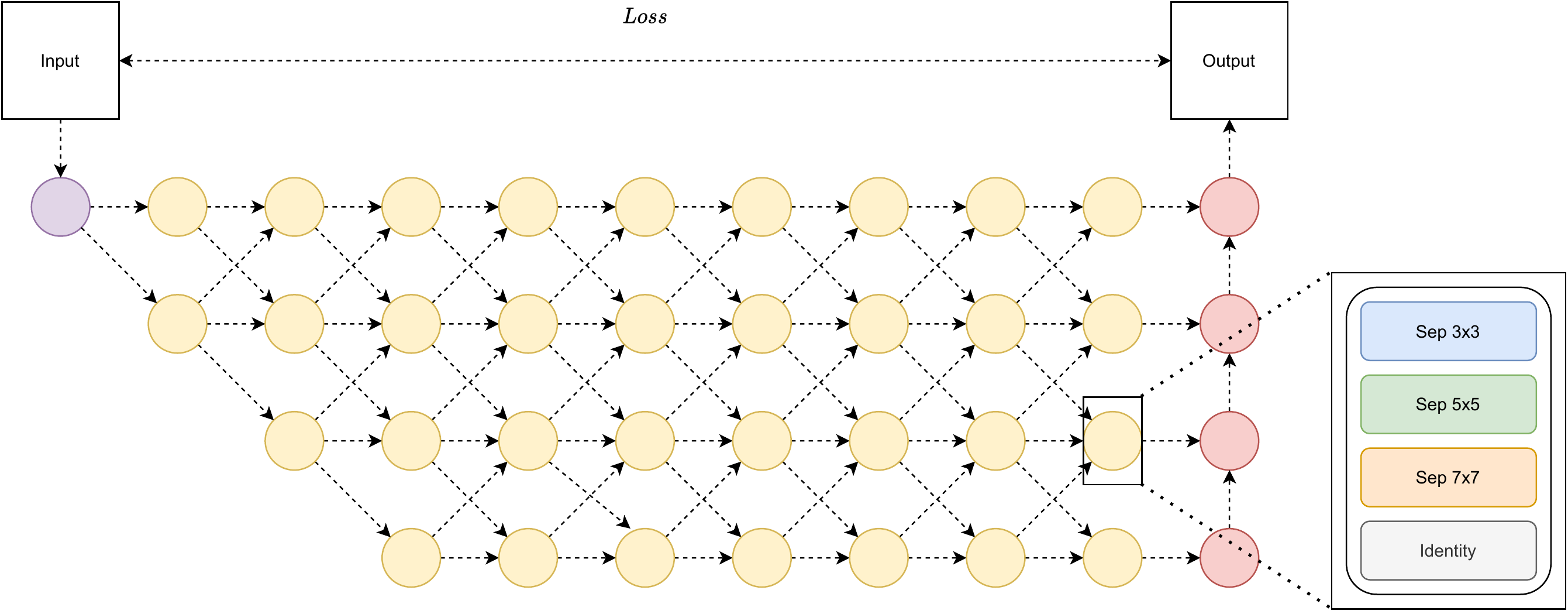}
    \caption{Diagram of the architecture used in our experiments. The purple, yellow and red dots represents respectively the "stem" convolution, the cells and the "upsampling aggregation" module. The arrows represent the data flow.}
    \label{fig:DN_schema}
\end{figure}

(\texttt{DR})
\label{sec:DN}
For the training phase, we employ the same structure and routing process as those proposed in \cite{li2020learning} (Figure \ref{fig:DN_schema}). The structure is composed of 4 down-sampling level, where the size of the features map is divided by $2$ at each level, but the depth of the latter is multiplied by $2$ using a  $1\times1$ 2D-convolution. In ours experiments we use a networks of $9$ layers, which correspond to $33$ cells in total (in yellow on Fig.\ref{fig:DN_schema}) . The structure also contains an \textit{"upsampling aggregation"} module at the end (red part on Fig.\ref{fig:DN_schema}). The goal of this module is to combine the features maps from all levels and reconstruct a map of the size of the input. Different from \cite{li2020learning}, here, each cells contains three \textit{transforming} operations (i.e. 2D-convolution with a kernel size of $3$, $5$ or $7$) to have a good point of comparison with the method described above. However, due to the structure of the network we decided not to use pooling operations, as this could have been potentially redundant. In addition, we have left the possibility of creating residual connections by adding operation identity in each cells. Moreover, we did not use the first two convolutions, originally used to reduce the size of the input, in order to keep as much information as possible. Instead, we used a $1\times1$ 2D-convolution (in purple on Fig.\ref{fig:DN_schema}). 

\subsection{Data-augmentation and Self-Learning phases}
\label{sec:LA}

After the search phase, the found \NN{} model with parameters $\theta$, $f_\theta$ is trained on $(X_{\ell},Y_{\ell})$ by minimizing the following loss~:
\begin{equation}
\label{eq:Loss1}
    \mathcal{L}(X_{\ell},Y_{\ell},\theta)=\ell(f_{\theta}(X_{\ell}),Y_{\ell})
    + \lambda \|\theta\|_2^2
    + \mu \Omega(f_{\theta}(X_{\ell}))
\end{equation}
where $\ell(.)$ is the Mean Absolute Error (\MAE{}),  and $\Omega(f_{\theta}(X_{\ell}))$ is the total variation function defined as:\\
\begin{equation*}
    \Omega(Z) = \sum_{i,j} \mid z_{i+1,j} - z_{i,j} \mid + \mid z_{i,j+1} - z_{i,j} \mid,
\end{equation*}\label{eq:loss_function}
with $z_{i,j}$ the measurement value of a point of coordinates $i,j$ in some signal distribution map $Z$. This function estimates the local amplitude variations of points in $Z$ that is minimized in order to ensure that neighbour points will have fairly close predicted measurements (i.e., preserving signal continuity/smoothness). 
Here, $\lambda$ and $\mu$ are hyperparameters for respectively the regularization and the total variation terms and they are found by cross-validation.
%Here, $\mu$ is the hyperparameter for the total variation term.
%

With Dynamic routing used in the search phase,  we optimize the network structure and the learning of parameters minimizing (Eq. \ref{eq:Loss1}) at the same time. Referring to Algorithm \ref{algo:nas-transductive}, the step 1 and 2 are combined in this case.
 
 Let $\theta^\star_1$ be the parameters of the optimized \NN{} model found by minimizing the loss \eqref{eq:Loss1} on ground truth measurements $(X_\ell,Y_\ell)$. This model is then applied to randomly chosen points, $X_u^{(k)}$, in $X_u$ and pseudo-\RSSI{} measurements $\tilde{Y}_u^{(k)}$ are obtained from the predictions of the optimized \NN{} model $f_{\theta^\star_1}$: $\tilde{Y}_u^{(k)} =~ f_{\theta^\star_1}(X_u^{(k)})$.   

With the same \NN{} architecture, a second model $f_{\theta^\star_2}$ is obtained by minimizing the loss \eqref{eq:Loss1} over the augmented training set $(X_{\ell},Y_{\ell})\cup (X_u^{(k)},\tilde{Y}_u^{(k)})$. %The pseudo-code of the proposed \SNAS{} approach is shown in Algorithm \ref{algo:nas-transductive}. 

\section{Experiments}
\label{sec:models}
% In this section the methods will be described.
% Radial basis function and kriging are the most known mathematical interpolation methods, that are used in the map restoration and function reconstruction. 
In this section we will first describe our experimental setup and then  present our experimental results.

\smallskip

\noindent \textbf{Experimental Setup.}
\label{subsec:datasets}
In all experiments, we considered maps of size $368 \times 368$ cells and tested our algorithm on field data from two distinct urban environments, namely the cities of Grenoble (France) and Antwerp (The Netherlands). We aggregated and averaged the given measurements in cells/pixels of size 10 meters x 10 meters. The Antwerp dataset is described in detail in \cite{antw_dataset} on which we considered three base stations, $BS'_1$, $BS'_2$ and $BS'_3$, with respectively $5969$,  $6450$ and $7118$ ground-truth measurements.
For the Grenoble dataset, we collected %two base stations in urban environment, for which we had amount of available 
GPS-tagged LoRa \RSSI{} measurements
with respect to 2 base stations located in different sites $BS_1$ and $BS_2$ with respectively  $16577$ and $7078$ ground truth measurements.
%The amount of available ground-truth measurements in the whole area of interest (Full map), around the base station (i.e., within a zoomed area) and %then amount of points 
To perform in-cell data aggregation, we measured the distances based on local East, North, Up (ENU) coordinates. %(in \texttt{python} library \texttt{pymap3d}), 
Then in each cell, we also computed the mean received power over all in-cell measurements (once converted into \RSSI{} values), before feeding our algorithm and the averaged \RSSI{} values have been normalized between 0 and~1.%The total amount of exploitable ground-truth data in the area of interest (located around the base station) are presented in Table \ref{tab:data}.

For each base station, 8\% of the pixels with ground-truth measurements were chosen for training $(X_\ell, Y_\ell)$, 2\% for validation $(X_v, Y_v)$ and the remaining 90\% for testing $(X_t, Y_t)$.The unlabeled data used in  \textbf{Step 3} of Algorithm \ref{algo:nas-transductive}  were selected at random from the remaining 4\% of each map's cells with no ground truth measurements.   Results are evaluated over the test set using the
\MAE{}, dB, estimated after re-scaling the normalized values to the natural received signal strength ones. The reported errors are averaged over $20$ random sets (training/validation/test) of the initial ground-truth data and unlabeled data were randomly chosen for each experiment. 

We compare  Radial basis functions (\RBF{}) \cite{bishop2007} with linear kernel that were found the most performant, kriging (\KRIG{}) \cite{oliver1990kriging}, and Navier-Stocks (\NAV) \cite{bertalmio2001navier} state-of-the-art interpolation techniques with the proposed \SNAS{} approach. For the latter, we employ both search phase methods based on Genetic Algorithm (\texttt{GA}) and 
Dynamic Routing (\texttt{DR}) and respectively referred to as \SNAS{}-\texttt{GA} and \SNAS{}-\texttt{DR}. For \SNAS{}-\texttt{GA} we also evaluate the impact of the self-training step (\textbf{Step 3}) (called \SNAS{}-\texttt{GA}($f_{\theta^\star_2}$)) by comparing it with the \NN{} model found at \textbf{Step 1} (called \SNAS{}-\texttt{GA}($f_{\theta^\star_1}$)).  The evolutionary algorithm in the architecture search phase (Section \ref{sec:ASP}) was implemented using the \texttt{NAS-DIP} 
\cite{ho2020nas_dip} package\footnote{\url{https://github.com/Pol22/NAS_DIP}}. The latter was developed over the Deep Image Prior (\DIP) method \cite{dip} which is a \NN{} model  proposed for image reconstruction. By considering \RSSI{} maps as corrupted images with partially observed pixels (ground-truth measurements), we also compare with this technique by training a \NN{} model having the same architecture than the one presented in \cite{dip} and referred to as \DIP{} in the following.
% \loic{Concerning the dynamic routing experiments, we follow the same procedure as described previously.} 
All experiments were run on Tesla V100 GPU.

%This program is based on the work of  
\smallskip

%As an unsupervised method, DIP~\cite{dip} uses an exact architecture of generative neural network, as a prior to the image for each of the tasks (in the original paper, they do super resolution, inpainting, denoising and image reconstruction). The network does not need external dataset for training, only the structure of the generative network itself to complete corrupted image. For each exact image, the algorithm finds the best parameters of the network for observed values of the pixels
\begin{figure}[t!]
    \begin{floatrow}
\capbtabbox{
\begin{scriptsize}
\begin{tabular}{c c c c c c c} \hline
         & \multicolumn{2}{c}{Grenoble} & &\multicolumn{3}{c}{Antwerp} \\%&  \\
                  & $BS_1$ & $BS_2$& & $BS'_1$ & $BS'_2$ & $BS'_3$  \\ \cline{2-3}\cline{5-7}
\RBF{} \cite{bishop2007}              & 5.03$^\downarrow$   & 3.16$^\downarrow$  & &3.58$^\downarrow$    & 3.35       &   3.90  \\% & 5 s  \\
\KRIG{} \cite{oliver1990kriging}                              & 5.68$^\downarrow$   &  4.21$^\downarrow$     & & 3.69$^\downarrow$       &  4.39$^\downarrow$    &  4.91$^\downarrow$   \\%& 2 min   \\
\NAV{} \cite{bertalmio2001navier}                                &  5.11$^\downarrow$      &  3.14$^\downarrow$     & & 4.28$^\downarrow$  &  3.45      &  3.87  \\%& 6 s    \\ 
\TV{}          &   5.13$^\downarrow$   &  2.89   & &   3.76   &  3.51    &    3.83    \\%& 3 min \\
\DIP{} \cite{dip}                     & 5.14$^\downarrow$ &  3.22$^\downarrow$   & &  3.53      &   3.41     &   3.92  \\%&  3 min  \\ 
\SNAS-\DR  &  4.82  & 2.82   & &  3.48    &   3.42     & 3.81   \\%& 1 h  \\
\SNAS-\texttt{GA}($f_{\theta^\star_1}$)         &  4.79  & 2.81  & & 3.39   &   \textbf{3.27}   &   3.75   \\%& 32 h  \\
\SNAS-\texttt{GA}($f_{\theta^\star_2}$)       &  \textbf{4.76}  &  \textbf{2.79}     & & \textbf{3.33}   & \textbf{3.27}    &   \textbf{3.74}  \\%& 32 h  \\
%\SNAS $(f_{\theta^\star_1})-cut$          &  4.61  & 2.81  & & 3.36   & 3.28    &   3.66   & 32 h  \\
%\SNAS $(f_{\theta^\star_2})-cut$          &  \textbf{4.60}  &  \textbf{2.79}     & &  3.35    &  \textbf{3.27}      &    \textbf{3.66}  & 32 h  \\
% \SNAS $(f_{\theta^\star_1})-gen$          &  4.73  & 2.81  & & 3.39   & 3.23    &   3.64    \\
% \SNAS $(f_{\theta^\star_2})-gen$          &  4.71  &  2.80     & &  3.35    &    3.22    &  3.64    \\

% $(f_{\theta^\star_1})$          &    &   & &    &     &       \\
% $(f_{\theta^\star_2})$          &    &       & &      &        &      \\
 \hline
    \end{tabular}
\end{scriptsize}
}{%
    \caption{Average values of the \MAE{}, dB of different approaches on all base stations. 
    % For the time indicator, \DR is not applicable (N/A) because several steps are performed simultaneously (See \textbf{Experimental results} paragraph).
    }\label{tab:results}
}
\ffigbox{
\centering
    \includegraphics[width=0.5\textwidth]{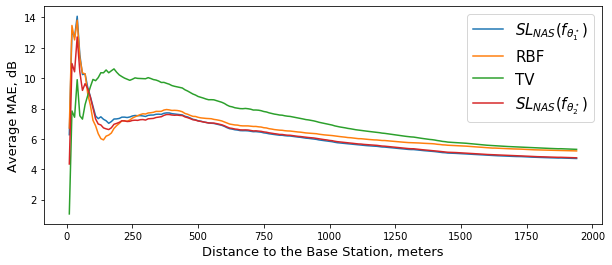}
 }{   \caption{MAE vs distance to the base station, $BS_1$.}
    \label{fig:MAE_vs_radius}
}
\end{floatrow}
\end{figure}

\noindent \textbf{Experimental Results.} 
\label{subsec:results}
Table \ref{tab:results} summarizes results obtained on the five considered \RSSI{} maps.
We use boldface to indicate the lowest errors. The symbol $^\downarrow$ indicates that the error is significantly higher than the best result with respect to Wilcoxon rank sum test used at a $p$-value threshold of $0.01$ \cite{wilcoxon45}. In all cases, \SNAS{}-\texttt{GA} and \SNAS{}-\texttt{DR} perform better than other state-of-the-art models even without the data-augmentation and self-training steps (\SNAS-\texttt{GA}($f_{\theta^\star_1}$)). We notice that \DIP{} which is also a \NN{} based model but with a fixed architecture has similar results than \RBF{}. These results show that the search of an optimized \NN{} model is effective for \RSSI{} map reconstruction in a constrained low-cost and low-complexity \IOT{} context.

Figure \ref{fig:MAE_vs_radius} depicts the average \MAE{} in dB with respect to the distance to the Base Station $BS_1$ for the city of Grenoble. For a distance above 250m, \SNAS{}-\texttt{GA}($f_{\theta^\star_2}$) provides uniformly better predictions in terms of \MAE{}. These results suggest further investigations for a better analysis of the model's predictions with respect to signal dynamics in regions where the signal is more erratic and where the dynamics is high in the case where additional contextual knowledge about the physical environment can be added in the learning process (e.g., typically as a side information channel or the city map).

\begin{wrapfigure}[21]{r}{0.6\textwidth}    \begin{center}
 \vspace{-9mm}   \begin{tikzpicture}[scale=.8]
\begin{axis}
            [
                ylabel = {\MAE{}, dB},
                boxplot/draw direction=y,
                xtick={1,2,3,4,5,6},
                xticklabels={\DIP, \RBF, 4\%, 7\%, 10\%, 14\%},
                x tick label style={font=\footnotesize, rotate=45,anchor=east}
            ]
    \addplot+[mark = *, 
    boxplot prepared={
        lower whisker=5.05,
        lower quartile=5.09,
        median=5.14,
        average = 5.14,
        upper quartile=5.18,
        upper whisker=5.31
    }, color = black
    ]coordinates{};% coordinates{(0,1.31)(0,1.32)(0,1.46)};
       \addplot+[mark = *, 
    boxplot prepared={
      lower whisker=4.96,
      lower quartile=4.98,
      median=5.04,
      average = 5.03,
      upper quartile=5.07,
      upper whisker=5.15
    }, color = black
    ]coordinates{};%oordinates{(0,0.91)(0,0.92)(0,1.85)};
   \addplot+[mark = *, 
      boxplot prepared={
      lower whisker=4.67,
      lower quartile=4.74,
      median=4.75,
      average = 4.76,
      upper quartile=4.78,
      upper whisker=4.79
    }, color = black
    ]coordinates{};% coordinates{(0,2.6)(0,2.67)(0,2.77)};
        \addplot+[mark = *, 
    boxplot prepared={
      lower whisker=4.69,
      lower quartile=4.73,
      median=4.77,
      average = 4.77,
      upper quartile=4.79,
      upper whisker=4.83
    }, color = black
    ]coordinates{};% coordinates{(0,1.71)(0,1.72)(0,3.22)};
    \addplot+[mark = *,
      boxplot prepared={
      lower whisker=4.7,
      lower quartile=4.74,
      median= 4.75,
      average = 4.79,
      upper quartile=4.78,
      upper whisker=4.84
    }, color = black
    ]coordinates{};
    \addplot+[mark = *, 
      boxplot prepared={
      lower whisker=4.68,
      lower quartile=4.72,
      median= 4.75,
      average = 4.78,
      upper quartile=4.79,
      upper whisker=4.86
    }, color = black
    ]coordinates{};

        \end{axis}
            \draw [decorate,decoration={brace,amplitude=5pt,mirror,raise=4ex}]
  (2.5,0) -- (6,0) node[midway,yshift=-3em]{\SNAS-\texttt{GA}($f_{\theta^\star_2}$)};
\end{tikzpicture}

\end{center}

    \caption{Boxplots showing the \MAE, dB distributions of \DIP{}, \RBF{}  and \SNAS-\texttt{GA} $(f_{\theta^\star_2})$ on $BS_1$ for different percentage of unlabeled data $\{4,7,10,14\}$ used in the self-learning phase. 
    }
    \label{fig:10perc_boxplot}
\end{wrapfigure}
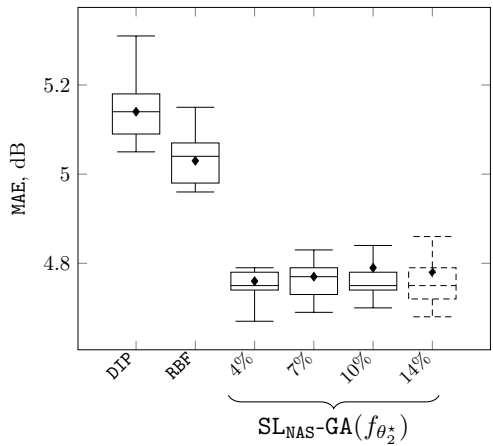

Figure \ref{fig:10perc_boxplot} displays the \MAE, dB boxplots
of  \DIP{}, \RBF{}  and \SNAS-\texttt{GA} $(f_{\theta^\star_2})$ on $BS_1$ for different percentages of unlabeled data used in the self-learning phase (Section \ref{sec:LA}). We notice that by increasing the size of unlabeled examples, the variance of \MAE{} for \SNAS-\texttt{GA}($f_{\theta^\star_2}$) increases mostly due to the increase of noisy predicted signal values by $f_{\theta^\star_1}$. This is mostly related to learning with imperfect supervisor that has been studied in semi-supervised learning \cite{NIPS2008_dc6a7071,Krithara:08}. As future work, we plan to incorporate a probabilistic label-noise model in \textbf{step 3} of algorithm \ref{algo:nas-transductive} and to learn simultaneously the parameters of the \NN{} and the label-noise models.

% \begin{figure}[hbt]
%     \centering
%     \includegraphics[width=0.6\linewidth]{images/barpl_10perc_1.png}
%     \caption{Percentage improvement in metrics compared to the baseline \RBF{}, 0.5x, 0.7x, etc. are the added amount of pixels in Split 2, where x is equal to 10\% of overall amount of points: train + validation}
%     \label{fig:10perc_barblot}
% \end{figure}

%

\section{Conclusion}
% In this paper, we proposed a first \NN{} based approach for reconstruction of received signal strength maps in a low-constraint and low-complexity context. The approach is based on the search of an optimal \NN{} architecture for the prediction of missing signal values in a \RSS{} map where we have only a small amount of ground-truth measurements, and, on self-learning.  

In this article, we presented a Neural Network model based on \NAS{} and self-learning for \RSS{}  map  reconstruction   from sparse single-snapshot input measurements in the absence of data augmentation via side deterministic simulations. We presented two variants for the search phase of \NAS{} based on Genetic algorithm and Dynamic routing. Experimental results on five large-scale maps of \RSS{} measurements reveal that our approach outperforms non-learning based interpolation state-of-the-art techniques and \NN{} with manually designed architecture.

%In this paper we proposed the self-learning algorithm based on search of the \NN{} architecture for the prediction of pixel values of the images and applied this method to the prediction of missing signal values in a \RSS{} map where we have only a small amount of ground-truth measurements. %\loic{Many directions are possible for future steps, including the tunning of the various hyper-parameters of models. } 

\bibliographystyle{splncs04}
\bibliography{refs}

\begin{thebibliography}{10}
\providecommand{\url}[1]{\texttt{#1}}
\providecommand{\urlprefix}{URL }
\providecommand{\doi}[1]{https://doi.org/#1}

\bibitem{antw_dataset}
Aernouts, M., Berkvens, R., Vlaenderen, K.V., Weyn, M.: Sigfox and lorawan
  datasets for fingerprint localization in large urban and rural areas. Data
  \textbf{3}(2) (2018)

\bibitem{NIPS2008_dc6a7071}
Amini, M.R., Usunier, N., Laviolette, F.: A transductive bound for the voted
  classifier with an application to semi-supervised learning. In: Advances in
  Neural Information Processing Systems. pp. 65--72 (2009)

\bibitem{bertalmio2001navier}
Bertalmio, M., Bertozzi, A.L., Sapiro, G.: Navier-stokes, fluid dynamics, and
  image and video inpainting. In: CVPR (2001)

\bibitem{bishop2007}
Bishop, C.M.: Pattern Recognition and Machine Learning (Information Science and
  Statistics). Springer-Verlag, Berlin, Heidelberg (2006)

\bibitem{burghal2020comprehensive}
Burghal, D., Ravi, A.T., Rao, V., Alghafis, A.A., Molisch, A.F.: A
  comprehensive survey of machine learning based localization with wireless
  signals (2020)

\bibitem{Cheng12}
Cheng, L., Wu, C., Zhang, Y., Wu, H., Li, M., Maple, C.: A survey of
  localization in wireless sensor network. International Journal of Distributed
  Sensor Networks  \textbf{2012} (12 2012)

\bibitem{LoRa-fingerprint_outdoor}
Choi, W., Chang, Y.S., Jung, Y., Song, J.: Low-power lora signal-based outdoor
  positioning using fingerprint algorithm. ISPRS International Journal of
  Geo-Information  \textbf{7}(11) (2018)

\bibitem{book_wireless_sensor_networks}
Dargie, W., Poellabauer, C.: Fundamentals of Wireless Sensor Networks: Theory
  and Practice. John Wiley \& Sons (2010)

\bibitem{radio_map_interpol_rbf}
Enrico, A., Redondi, C.: {Radio Map Interpolation Using Graph Signal
  Processing}. IEEE Communications Letters  \textbf{22}(1),  153--156 (2018)

\bibitem{SVT_RSSI_crowdsensing}
Fan, X., He, X., Xiang, C., Puthal, D., Gong, L., Nanda, P., Fang, G.: Towards
  system implementation and data analysis for crowdsensing based outdoor {RSS}
  maps. {IEEE} Access  \textbf{6},  47535--47545 (2018)

\bibitem{ho2020nas_dip}
Ho, K., Gilbert, A., Jin, H., Collomosse, J.: Neural architecture search for
  deep image prior (2020)

\bibitem{nas_bo}
Jin, H., Song, Q., Hu, X.: Auto-keras: An efficient neural architecture search
  system. In: Proceedings of the 25th ACM SIGKDD. pp. 1946--1956 (2019)

\bibitem{localization_iot_survey}
Khelifi, F., Bradai, A., Benslimane, A., Rawat, P., Atri, M.: {A Survey of
  Localization Systems in Internet of Things}. {Mobile Networks and
  Applications}  \textbf{24}(3),  761--785 (Jun 2019)

\bibitem{Krithara:08}
Krithara, A., Amini, M.R., Renders, J.M., Goutte, C.: Semi-supervised document
  classification with a mislabeling error model. In: 30th European Conference
  on Information Retrieval. pp. 370--381. Glasgow (2008)

\bibitem{path_loss_wifi_indoor}
{Kubota}, R., {Tagashira}, S., {Arakawa}, Y., {Kitasuka}, T., {Fukuda}, A.:
  Efficient survey database construction using location fingerprinting
  interpolation. In: 2013 IEEE 27th International Conference on Advanced
  Information Networking and Applications (AINA). pp. 469--476 (2013)

\bibitem{Laaraiedh1}
{Laaraiedh}, M., {Uguen}, B., {Stephan}, J., {Corre}, Y., {Lostanlen}, Y.,
  {Raspopoulos}, M., {Stavrou}, S.: Ray tracing-based radio propagation
  modeling for indoor localization purposes. In: 2012 IEEE 17th International
  Workshop on Computer Aided Modeling and Design of Communication Links and
  Networks (CAMAD). pp. 276--280 (2012)

\bibitem{RadioUNet}
{Levie}, R., {Yapar}, V., {Kutyniok}, G., {Caire}, G.: Pathloss prediction
  using deep learning with applications to cellular optimization and efficient
  d2d link scheduling. In: {ICASSP}. pp. 8678--8682 (2020)

\bibitem{Li_heap_overview_interp}
Li, J., D.Heap, A.: A review of comparative studies of spatial interpolation
  methods in environmental sciences: Performance and impact factors. Ecological
  Informatics  \textbf{6}(3),  228 -- 241 (2011)

\bibitem{li2020learning}
{Li}, Y., {Song}, L., {Chen}, Y., {Li}, Z., {Zhang}, X., {Wang}, X., {Sun}, J.:
  Learning dynamic routing for semantic segmentation (2020)

\bibitem{kriging_plus_crowdsensing}
Liao, J., Qi, Q., Sun, H., Wang, J.: Radio environment map construction by
  kriging algorithm based on mobile crowd sensing. Wireless Communications and
  Mobile Computing  \textbf{2019},  1--12 (02 2019)

\bibitem{autodeep}
Liu, C., Chen, L.C., Schroff, F., Adam, H., Hua, W., Yuille, A.L., Fei-Fei, L.:
  Auto-deeplab: Hierarchical neural architecture search for semantic image
  segmentation. In: Proceedings of {CVPR}. pp. 82--92 (2019)

\bibitem{darts}
Liu, H., Simonyan, K., Yang, Y.: Darts: Differentiable architecture search.
  arXiv preprint arXiv:1806.09055  (2018)

\bibitem{path_loss_rssi_outdoor_clustering}
{Ning}, C., et~al.: Outdoor location estimation using received signal
  strength-based fingerprinting. Wireless Pers Commun  \textbf{99},  365–384
  (2016)

\bibitem{oliver1990kriging}
Oliver, M., Webster, R.: Kriging: a method of interpolation for geographical
  information systems. International Journal of Geographical Information System
   \textbf{4}(3),  313--332 (1990)

\bibitem{Raspopoulos2}
{Raspopoulos}, M., {Laoudias}, C., {Kanaris}, L., {Kokkinis}, A., {Panayiotou},
  C.G., {Stavrou}, S.: 3d ray tracing for device-independent fingerprint-based
  positioning in wlans. In: 2012 9th Workshop on Positioning, Navigation and
  Communication. pp. 109--113 (2012)

\bibitem{regu_evo}
Real, E., Aggarwal, A., Huang, Y., Le, Q.V.: Regularized evolution for image
  classifier architecture search. In: {AAAI}. vol.~33, pp. 4780--4789 (2019)

\bibitem{rbf_interp}
{Redondi}, A.E.C.: Radio map interpolation using graph signal processing. IEEE
  Communications Letters  \textbf{22}(1),  153--156 (2018)

\bibitem{RFB15a}
Ronneberger, O., P.Fischer, Brox, T.: U-net: Convolutional networks for
  biomedical image segmentation. In: Medical Image Computing and
  Computer-Assisted Intervention (MICCAI). LNCS, vol.~9351, pp. 234--241.
  Springer (2015)

\bibitem{rbf_kriging}
Rusu, V., Rusu, C.: {Radial Basis Functions Versus Geostatistics in Spatial
  Interpolations}. vol.~217 (10 2006)

\bibitem{kriging_ffnn_pl}
{Sato}, K., {Inage}, K., {Fujii}, T.: On the performance of neural network
  residual kriging in radio environment mapping. IEEE Access  \textbf{7},
  94557--94568 (2019)

\bibitem{Sorour2}
{Sorour}, S., {Lostanlen}, Y., {Valaee}, S., {Majeed}, K.: Joint indoor
  localization and radio map construction with limited deployment load. IEEE
  Transactions on Mobile Computing  \textbf{14}(5),  1031--1043 (2015)

\bibitem{Tahat16}
{Tahat}, A., {Kaddoum}, G., {Yousefi}, S., {Valaee}, S., {Gagnon}, F.: A look
  at the recent wireless positioning techniques with a focus on algorithms for
  moving receivers. IEEE Access  \textbf{4},  6652--6680 (2016)

\bibitem{dip}
Ulyanov, D., Vedaldi, A., Lempitsky, V.: Deep image prior. CoRR
  \textbf{abs/1711.10925} (2017)

\bibitem{Vo16}
{Vo}, Q.D., {De}, P.: {A Survey of Fingerprint-Based Outdoor Localization}.
  IEEE Communications Surveys \& Tutorials  \textbf{18}(1),  491--506 (2016)

\bibitem{wilcoxon45}
Wilcoxon, F.: Individual comparisons by ranking methods. Biometrics
  \textbf{1}(6),  80--83 (1945)

\bibitem{Yu09}
Yu, K., Sharp, I., Guo, Y.: Ground-Based Wireless Positioning. John Wiley \&
  Sons, Ltd (06 2009)

\bibitem{CEDGAN}
Zhu, D., Cheng, X., Zhang, F., Yao, X., Gao, Y., Liu, Y.: Spatial interpolation
  using conditional generative adversarial neural networks. International
  Journal of Geographical Information Science  \textbf{34}(4),  735--758 (2020)

\bibitem{nas_rl}
Zoph, B., Le, Q.V.: Neural architecture search with reinforcement learning.
  arXiv preprint arXiv:1611.01578  (2016)

\end{thebibliography}

\end{document}